\documentclass[10pt,twocolumn,letterpaper]{article}

\usepackage[pagenumbers]{wacv} 

\usepackage{xcolor}
\definecolor{darkgreen}{rgb}{0.0,0.5,0.0}
\definecolor{darkred}{rgb}{0.55,0.0,0.0}
\usepackage[table]{xcolor}
\usepackage{multirow}
\usepackage{caption}
\usepackage{siunitx}   
\usepackage{colortbl}  
\usepackage{wrapfig}
\usepackage{xcolor}   
\usepackage{pifont} 
\usepackage{tabularx}
\usepackage{colortbl}   
\usepackage{pifont}     
\usepackage{rotating} 
\usepackage{graphicx}
\usepackage{booktabs}
\usepackage{multirow}
\usepackage{booktabs}
\usepackage{pifont}
\usepackage{xcolor,colortbl}
\usepackage{stfloats}
\newcommand{\cmark}{\ding{51}}
\newcommand{\xmark}{\ding{55}}
\definecolor{darkgreen}{RGB}{0,120,0}
\definecolor{darkred}{RGB}{150,0,0}
\colorlet{ourscol}{blue!8}
\newcommand{\rot}[1]{\rotatebox{90}{\hspace{1pt}#1}}

\definecolor{wacvblue}{rgb}{0.21,0.49,0.74}
\usepackage[pagebackref,breaklinks,colorlinks,allcolors=wacvblue]{hyperref}

\def\wacvPaperID{*****} 
\def\confName{WACV}
\def\confYear{2027}

\title{Reconstruct What You Can: Distractor-free 4D Gaussian Splatting}

\author{Joel Julin$^{1}$ \quad Liuyue Xie$^{1}$ \quad Ananya Bal$^{1}$\\
Mose Sakashita$^{2}$ \quad Koichiro Niinuma$^{2}$ \quad Laszlo A. Jeni$^{1}$\\
$^{1}$Carnegie Mellon University \qquad $^{2}$Fujitsu Research of America\\
{\small\url{https://recan-gs.github.io}}}

\begin{document}
\twocolumn[{%
    \renewcommand\twocolumn[1][]{#1}%
    \maketitle
    \vspace{-1.4cm}
    \begin{center}
        \includegraphics[width=\textwidth]{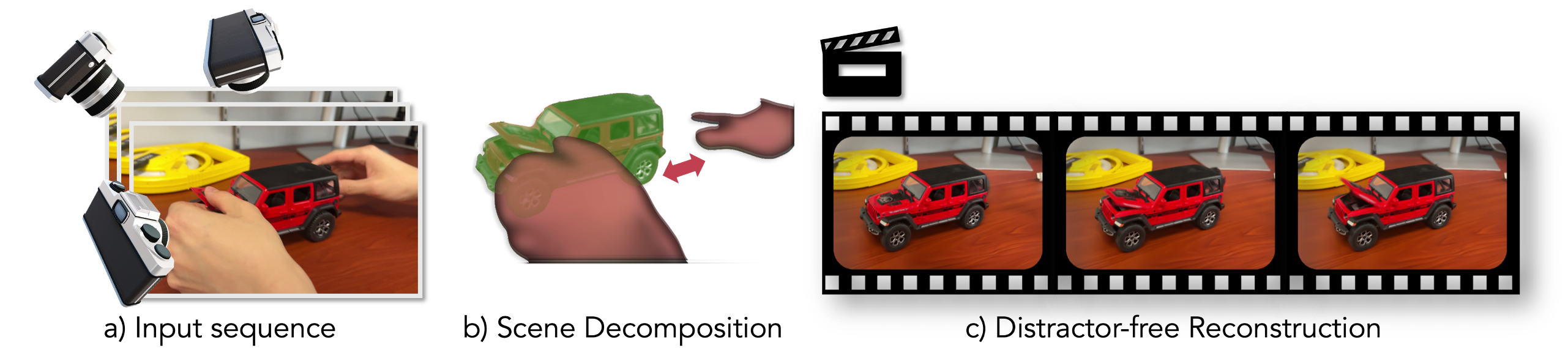}
        \captionof{figure}{
    From monocular frames of noisy, dynamic scenes (a), our method decomposes reconstructable \textcolor{darkgreen}{(green)} and unreconstructable motion \textcolor{darkred}{(red)} (b), enabling distractor-free dynamic reconstruction (c) while preserving fine-scale details.
}
        \label{fig:teaser}
    \end{center}
}]
\begin{abstract}
Dynamic reconstruction from casual video must preserve valid subject motion while suppressing transient occluders, shadows, and volumetric effects. Existing methods either suppress valid motion through static-scene assumptions or model all observations jointly, producing ghosting and unstable geometry. We introduce ReCan-GS, a 4D Gaussian Splatting framework that separates coherent scene content from inconsistent observations. It derives a Jacobian-based deformation-sensitivity score from the learned deformation field, using it as an empirical proxy for reconstruction ambiguity. The score guides separation into dynamic Core Gaussians and per-view Distractor Gaussians, or post-hoc pruning of sensitive Core Gaussians. The same criterion supports removal of spatial occluders, photometric artifacts, and evolving smoke or fog without masks, pretrained segmenters, or medium-specific supervision. Experiments on synthetic and real dynamic scenes and synthetic smoke/fog scenes demonstrate its usefulness across the evaluated distractor regimes.

\end{abstract}
    
\section{Introduction}
\label{sec:intro}
Reconstructing dynamic 3D scenes from unconstrained monocular video remains challenging because real-world sequences often contain transient occluders, reflections, shadows, and flickers. These effects can introduce conflicting constraints on a shared scene representation. Standard dynamic reconstruction methods attempt to model them jointly with the subject, which can produce ghosting and unstable geometry.

Static distractor-removal methods avoid this failure mode by treating dynamic content as noise, but therefore fail when the moving object is the reconstruction target. We instead define distractors by deformation-space instability rather than object identity, speed, or appearance. This lets reconstructable motion remain in the core scene while inconsistent content is routed away.

The same criterion also handles slow-moving volumetric distractors such as fog and smoke: although they may appear photometrically stable, their evolving appearance can induce inconsistent deformation constraints and high scores, enabling removal by post-hoc pruning without fog-specific design.

To this end, we present \textbf{ReCan-GS}, a 4D Gaussian Splatting framework for distractor removal in dynamic scenes. ReCan-GS derives a motion-stability uncertainty directly from the Jacobian of the learned deformation field. This signal is intended to highlight observations associated with sensitive or inconsistent deformation, allowing the model to route stable content into Core Gaussians and inconsistent content into a separate Per-View Distractor layer. ReCan-GS is primarily designed for dynamic scenes with transient distractors and
for volumetric fog and smoke removal; its performance is more limited on standard static
distractor benchmarks.

Our main contributions are:
\begin{itemize}
\item A Jacobian-derived uncertainty metric that quantifies deformation-space instability in a continuous 4D Gaussian deformation field, providing a routing cue without explicit appearance models or speed thresholds.

\item A unified distractor-removal framework for 4D Gaussian Splatting that separates geometrically consistent content from inconsistent observations, covering dynamic-scene occluders, photometric artifacts, and dynamic volumetric distractors such as fog and smoke.

\item Empirical validation across synthetic and real dynamic scenes with distractors, plus fog/smoke scenes with clean ground truth.
\end{itemize}

\section{Related Works}
\label{sec:related_work}
  
\subsection{Handling Distractors}
Recent research (Table \ref{tab:method_comparison}) addresses the challenge of transient distractors within predominantly static scene assumptions. Approaches generally fall into three categories based on their distractor-handling strategies:

\textbf{Photometric and Uncertainty-Based Methods} identify distractors through reconstruction error or predictive uncertainty. NeRF-W \cite{martinbrualla2020nerfw} first separated static and transient components. BayesRays \cite{goli2023} extended this with probabilistic uncertainty, while FisherRF \cite{Jiang2023FisherRF} used Fisher information. RobustNeRF \cite{Sabour_2023_CVPR} uses robust statistics to mask high-error regions. Related work addresses reflections and inconsistent visibility through appearance and regularization strategies~\cite{verbin2022refnerf, barron2021mipnerf, Niemeyer2021Regnerf, scaffoldgs}. Concurrent work \cite{guo2026uncertainty} estimates per-Gaussian reliability from recurring observations for motion propagation under occlusion; our score instead measures deformation-field sensitivity to separate transient content.

\textbf{Semantic Feature-Based Methods} replace photometric cues with semantic information from pretrained vision models. In the NeRF domain, NeRF-HuGS \cite{chen2024nerf} uses semantic segmentation, and NeRF On-the-go \cite{Ren2024NeRF} leverages DINOv2-derived \cite{oquab2023dinov2} features. This trend extends to 3DGS, where SpotLessSplats \cite{sabourgoli2024spotlesssplats} clusters text-to-image embeddings and Robust 3D Gaussian Splatting \cite{ungermann2024robust} incorporates SAM-derived segmentations \cite{kirillov2023segany}.

\begin{table}[t]
\centering
\scriptsize
\setlength{\tabcolsep}{2.5pt}
\renewcommand{\arraystretch}{1.05}
\colorlet{ourscol}{blue!8}
\begin{tabularx}{\columnwidth}{lXXXXXXXX}
\toprule
\textbf{Property} &
\rot{\textbf{SWAG}~\cite{dahmani2024swagsplattingwildimages}} &
\rot{\textbf{GS-W}~\cite{zhang2024GS-W}} &
\rot{\textbf{Wild-GS}~\cite{xu2024wildgsrealtimenovelview}} &
\rot{\textbf{Splatfacto-W}~\cite{xu2024splatfactownerfstudioimplementationgaussian}} &
\rot{\textbf{WildGaussians}~\cite{kulhanek2024wildgaussians}} &
\rot{\textbf{SpotlessSplats}~\cite{sabourgoli2024spotlesssplats}} &
\rot{\textbf{DeSplat}~\cite{wang2024desplat}} &
\cellcolor{ourscol}\rot{\textbf{Ours}} \\
\midrule
No pre-trained model          & \cmark & \xmark & \xmark & \cmark & \xmark & \cmark & \cmark & \cellcolor{ourscol}\cmark \\
Fast 3DGS rendering           & \xmark & \xmark & \xmark & \cmark & \cmark & \cmark & \cmark & \cellcolor{ourscol}\cmark \\
Explicit scene separation     & \xmark & \xmark & \xmark & \xmark & \xmark & \cmark & \cmark & \cellcolor{ourscol}\cmark \\
Dynamic scene reconstruction  & \xmark & \xmark & \xmark & \xmark & \xmark & \xmark & \xmark & \cellcolor{ourscol}\cmark \\
Uncertainty Metric    & \xmark & \xmark & \xmark & \xmark & \xmark & \xmark & \xmark & \cellcolor{ourscol}\cmark \\
Fog/Smoke Removal    & \xmark & \xmark & \xmark & \xmark & \xmark & \xmark & \xmark & \cellcolor{ourscol}\cmark \\
\bottomrule
\end{tabularx}
\vspace{8pt}
\caption{\textbf{Feature comparison.} Our method uniquely supports distractor removal in fully dynamic scenes without pretraining.}
\label{tab:method_comparison}
\vspace{-0.5cm}
\end{table}

\textbf{Decompositional Methods} explicitly separate static and transient content into distinct representational layers. Several methods model appearance variation using learned embeddings or fields (SWAG \cite{dahmani2024swagsplattingwildimages}, GS-W \cite{zhang2024GS-W}, Wild-GS \cite{xu2024wildgsrealtimenovelview}). Others use optimization strategies, such as WildGaussians \cite{kulhanek2024wildgaussians}, which uses uncertainty masks for gradient scaling.

A more explicit decomposition is used by DeSplat \cite{wang2024desplat}, which models transients with per-view splats, and HybridGS \cite{lin2024hybridgs}, which combines persistent 3D Gaussians with view-specific 2D Gaussians. Splatfacto-W \cite{xu2024splatfactownerfstudioimplementationgaussian} also uses a hybrid approach with MLPs for background modeling and robust masks for transients.


While these methods are effective in their respective contexts, they share a common limitation: all assume a static scene and ultimately aim to recover a single clean reconstruction, with uncertainty or transient handling largely heuristic or learned. In contrast, we address fully dynamic scenes corrupted by severe spatial and photometric inconsistencies and, importantly, show that uncertainty can be derived analytically, offering a principled and explainable view of the optimization process that prior work lacks.

\subsection{Fog and Volumetric Distractor Removal}
Volumetric distractors such as fog and smoke further challenge appearance-based removal, since they may appear photometrically stable while remaining geometrically inconsistent across views. Our method instead uses deformation sensitivity as a cue for removal without fog-specific supervision. The experiments evaluate this mechanism against general distractor-removal baselines; they do not establish competitiveness with dedicated smoke-removal methods.

\subsection{Dynamic Scene Reconstruction}

Reconstructing general dynamic scenes is a distinct but related challenge. Methods such as DeformableGS~\cite{yang2023deformable3dgs}, 4DGS~\cite{Wu_2024_CVPR}, GSLK~\cite{xie2025gaussiansplattinglucaskanade}, DynamicGS~\cite{luiten2023dynamic}, STGS~\cite{Li_STG_2024_CVPR}, Realtime4DGS~\cite{yang2023gs4d}, and dynamic NeRF variants~\cite{pumarola2020d, park2021nerfies, park2021hypernerf, kplanes_2023} model time-varying scenes by learning motion jointly with appearance. However, they lack an explicit mechanism to separate transient distractors from core scene motion. As a result, they tend to average all visual phenomena, leading to severe artifacts such as ghosting when occluders are present.

More recent approaches, such as Shape of Motion~\cite{som2024} and MoSca~\cite{lei2024mosca}, leverage strong externally-derived priors from vision foundation models (e.g., point tracking or depth estimation) to regularize the motion field. While effective, these methods introduce additional complexity and depend heavily on the quality of the pretrained models. In contrast, our approach is orthogonal: we propose an analytical, prior-free method that explicitly separates geometrically consistent motion from temporal multi-view contradictions using an uncertainty signal derived directly from the deformation field, improving both robustness and interpretability.

\section{Preliminary}
\label{sec:prelim}
3DGS~\cite{kerbl3Dgaussians} represents a scene as a set of 3D Gaussians, each
defined by a position $\mathbf{\mu}$, covariance $\mathbf{\Sigma}$, opacity
$\alpha$, and view-dependent color via spherical harmonics. Gaussians are
initialized from a sparse SfM point cloud~\cite{Micheletti2015StructurePhotogrammetry}
and adaptively split or pruned during training. Pixel colors are computed by
alpha-compositing depth-sorted 2D Gaussian projections:
\begin{equation}
    \mathbf{C} = \sum^{N}_{i=1}T_{i}\bigl(1-\exp(-\sigma_{i}\delta_{i})\bigr)\mathbf{c}_{i}, \quad
    T_{i} = \exp\!\Bigl(-\sum_{j=1}^{i-1}\sigma_{j}\delta_j\Bigr).
    \label{eq:render}
\end{equation}
For dynamic scenes, a common approach is to learn a deformation field mapping a
canonical representation to its configuration at a given
time~\cite{pumarola2020d, park2021nerfies, luiten2023dynamic, yu2023cogs}. We
adopt this strategy for Gaussian Splatting following~\cite{yang2023deformable3dgs},
using an MLP that maps each canonical position $\mu_i$ and time $t$ to position,
rotation, and scale offsets.





\section{Method}
\begin{figure*}[t]
    \centering
    \includegraphics[width=0.895\linewidth]{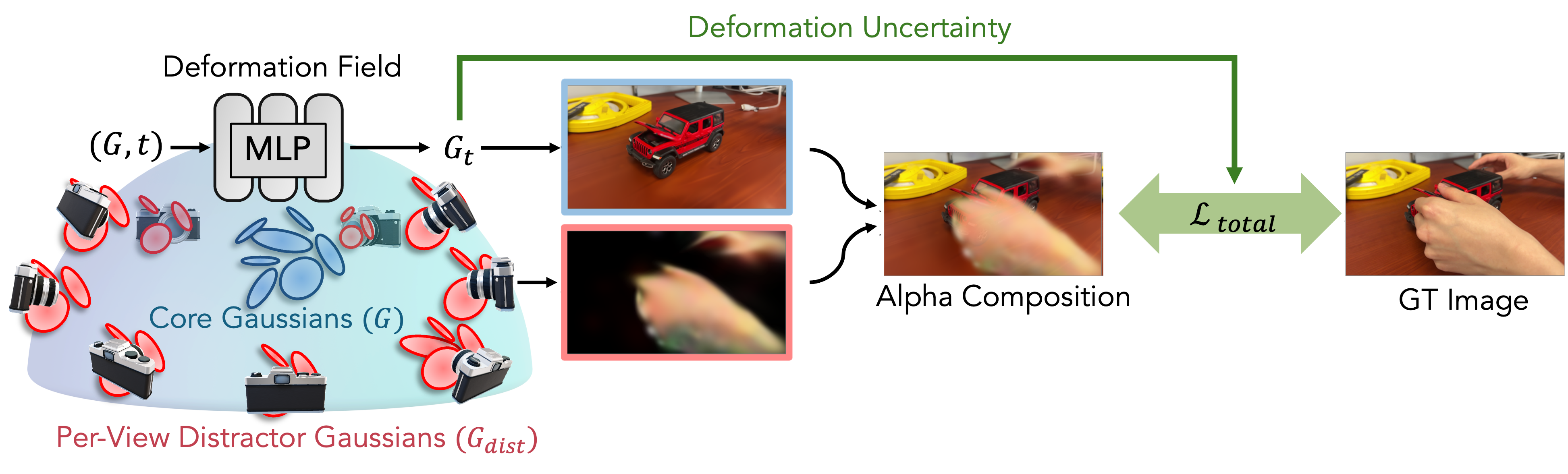}
    \caption{An overview of our ReCan-GS pipeline. The scene is modeled with two components: dynamic Core Gaussians ($G$) and per-view Distractor Gaussians ($G_{dist}$). An MLP deforms the core Gaussians over time ($t$). A Jacobian-based uncertainty signal, derived from the deformation field, guides the optimization to penalize unstable motion in the core layer, effectively forcing transient effects into the distractor layer.}
    \label{fig:pipeline}
\end{figure*}

To robustly reconstruct dynamic scenes with transient distractors, ReCan-GS
decomposes the scene into two geometric layers: dynamic Core Gaussians modeling
stable, temporally consistent content, and per-view Distractor Gaussians
capturing transient or view-specific phenomena. The separation is driven by a
motion-derived uncertainty signal from the learned deformation field, reinforced
by competition between the two layers during optimization, and requires no masks
or annotations. Fig.~\ref{fig:pipeline} shows the pipeline.
\vspace{-0.3cm}
\paragraph{Definition of Distractors.}
We operationally identify candidate distractors as observations that induce inconsistent deformation constraints in the shared canonical space. Routing uses deformation sensitivity rather than an explicit threshold on speed or object identity. This criterion does not determine the intended subject: a well-observed distractor can admit stable deformation, while poorly observed target content can receive a high score. The separation therefore depends on observation coverage and the learned representation. 

Sec.~\ref{sec:ablation} evaluates coverage and speed
(Fig.~\ref{fig:ablation}), pruning sensitivity (Fig.~\ref{fig:post_hoc}),
and distractor localization; smoke results provide a further test of the same
score (Fig.~\ref{fig:smoke_results}).

\subsection{Decomposed 4D Gaussian Splatting}
\label{decomp}

Our method represents the scene with two distinct sets of 3D Gaussians, drawing inspiration from decompositional ideas explored in static settings such as DeSplat~\cite{wang2024desplat}.

\paragraph{Core Gaussians.}
The first component, the Core Gaussians $G_{core}$
 , models the stable scene content, including the static background and the primary dynamic subject. These Gaussians are defined in a canonical space and made dynamic via a learned deformation field $N$, implemented as an 8-layer, 256-wide MLP. For a given canonical position $\mu_{i}$ and time t, the MLP predicts deformation offsets:

\begin{equation}
\label{def_net}
        N(\mu_i, t) = (\Delta\mu_i, \Delta q_i, \Delta s_i)
\end{equation}

These offsets update the position, rotation, and scale of each Gaussian to reflect its configuration at time t, enabling complex, non-rigid motion. This MLP architecture is identical to \cite{yang2023deformable3dgs}.

\vspace{-0.3cm}

\paragraph{Distractor Gaussians.}
We adopt the per-view distractor formulation of DeSplat~\cite{wang2024desplat}, which provides a simple mechanism for modeling view-dependent transients. For completeness, we summarize the components here, noting that our primary contribution lies in the uncertainty-guided routing described in Sec.~\ref{sec: guiding_w_dyn}.

\textit{Initialization.} Each training view n maintains its own distractor set $G_{dist,n}$. We initialize K Gaussians on a shallow camera-aligned plane via:
\begin{equation}
    \boldsymbol{\mu}_d = \mathbf{t}_n - R_n^\top \mathbf{u},
    \qquad
    \mathbf{u} =
    \begin{bmatrix}
        \rho u \\ \rho v \\ \rho
    \end{bmatrix},
    \quad
    u,v \sim \mathcal{U}(0,1),
\end{equation}
where $\rho$ controls the plane depth. Colors are parameterized as RGB vectors to reflect their view-dependent nature.

\textit{Densification.} Distractor Gaussians undergo per-view Adaptive Density Control based on view-space gradients. Updates occur only when a per-view counter $s_{n}$ exceeds threshold $S$, allowing each view to refine its own distractor set independently.

\vspace{-0.3cm}
\paragraph{Compositing.}
The core and distractor layers are rendered separately and then combined via standard alpha compositing:

\begin{equation}
    c_{\text{comp}}
    = c_{\text{dist}} + (1 - \alpha_{\text{dist}})\, c_{\text{core}}.
\end{equation}

This enables distractors to occlude the core where appropriate.

\subsection{Guiding Decomposition with Scene Dynamics}
\label{sec: guiding_w_dyn}
A standard photometric loss is insufficient to enforce a meaningful separation, as it encourages the powerful Core model \(G_{\text{core}}\) to overfit to all visual phenomena. To resolve this, we leverage a Jacobian-derived uncertainty score to actively guide the decomposition, identifying geometrically inconsistent motion for routing to $G_{\text{dist}}$ or removal.

\paragraph{Jacobian Derived Motion Uncertainty.}
Our guiding signal is derived from the core deformation field $N$ and measures variation among its output gradients. We use this deformation-sensitivity score as an empirical proxy for reconstruction ambiguity, referring to it as Jacobian uncertainty. It is not a calibrated probabilistic uncertainty estimate: complex but correctly reconstructed deformation may also yield a high score.

In the controlled experiment of Sec.~\ref{sec:ablation}, increasing camera coverage reduces the score at both tested motion speeds. This supports its usefulness beyond raw velocity, but does not guarantee that valid motion will always receive low uncertainty.

For each Core Gaussian $g_i$, we compute the Jacobian of the predicted offsets $(\Delta\mu_i, \Delta q_i, \Delta s_i)$ with respect to the inputs (canonical position $\mu_i$ and time $t$). For example, for the position offset $\Delta\mu$, the Jacobian $J_{\Delta\mu}$ is the matrix of partial derivatives:
\begin{equation}
J_{\Delta\mu} = \frac{\partial \Delta\mu}{\partial (\mu, t)}
\end{equation}

Our uncertainty score is derived from the sample covariance of the Jacobian's rows, which represent the gradients of each output dimension with respect to the inputs. High variance among these gradients indicates heterogeneous local sensitivity, which we use as a routing cue. The uncertainty contribution from a single parameter type (e.g., position) is the trace of this covariance matrix $U_{\Delta\mu}(g_i) = \text{Tr}\left(\text{Cov}(J_{\Delta\mu})\right)$.

The trace represents the sum of the variances of its constituent variables, providing a unified measure of deformation sensitivity. In practice, we compute this Jacobian using a stochastically perturbed time input $t' = t + \epsilon$ for a robust local estimate. The total uncertainty score for a Gaussian $g_i$ is the sum of the scores for each parameter type:
\begin{equation}
U(g_i) = U_{\Delta\mu}(g_i) + U_{\Delta q}(g_i) + U_{\Delta s}(g_i)
\end{equation}

\paragraph{Uncertainty-Guided Methods}
Once high-uncertainty regions are detected, we use one of two methods to suppress their influence in the core model and encourage a handover to the distractor model.

\textit{Variant 1: Loss-Based Regularization.} Our primary approach integrates the uncertainty score directly into the optimization process. We adopt the alpha-based regularization from DeSplat~\cite{wang2024desplat}, which encourages a sparse distractor layer and an opaque core layer ($ \mathcal{L}_{\text{alpha}} = \lambda_{s}|1-\alpha_{\text{core}}| + \lambda_{d}|\alpha_{\text{dist}}|$), we introduce an uncertainty-aware regularizer:

\begin{equation}
    \mathcal{L}_{\text{unc}} = \sum_{g_{i} \in G_{\text{core}}} U(g_{i}) \cdot \alpha_{\text{core}, i}
\end{equation}
This term penalizes core Gaussians that are both uncertain and opaque, encouraging uncertain regions to transfer to the distractor layer.  
The final loss function combines these terms:
\begin{equation}
\label{eq:opt}
    \mathcal{L}_{total} = \mathcal{L}_{photo} + \mathcal{L}_{alpha} + \lambda_{unc}\mathcal{L}_{unc}
\end{equation} 

\textit{Variant 2: Post-Hoc Uncertainty Pruning.}
For severe clutter, we also consider a hard post-training separation. In this approach, we omit the uncertainty regularizer $\mathcal{L}_{\text{unc}}$ during training and instead track the stability of each core Gaussian throughout optimization. We accumulate uncertainty using an exponential moving average (EMA):
\begin{equation}
    U_{t}(g_i) = \beta\, U_{t-1}(g_i) + (1 - \beta)\, U_{\text{inst}}(g_i),
\end{equation}
where $U_{\text{inst}}(g_i)$ is the instantaneous uncertainty at iteration $t$ and $\beta$ is a momentum parameter. After training converges, any Core Gaussian whose accumulated uncertainty exceeds a threshold is removed:
\begin{equation}
    g_i \in \text{top-}r\%\{U_t(g_j)\}_{j}.
\end{equation}
This pruning step eliminates residual unstable geometry that may remain when relying solely on soft regularization. As demonstrated in Fig.~\ref{fig:post_hoc}, adjusting the pruning threshold hierarchically isolates topological artifacts, ranging from severe spatial occluders to subtle photometric inconsistencies (evaluated in Sec.~\ref{sec:ablation}).


\section{Experiments}

\begin{figure*}[t]
\centering
\includegraphics[width=0.81\linewidth]{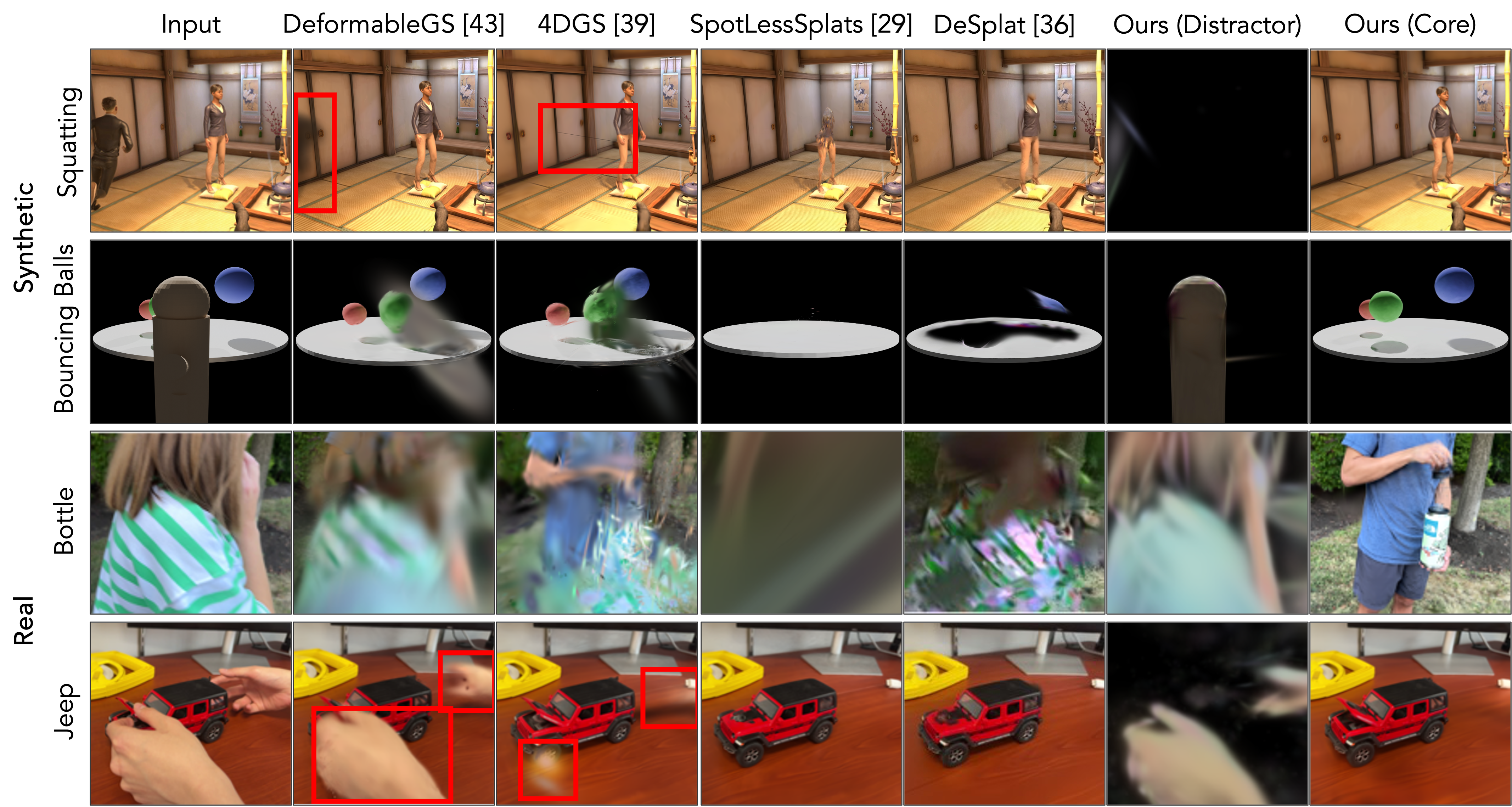}
\caption{\textbf{Qualitative results on dynamic scenes with distractors.} Columns 2–3 are dynamic reconstruction methods that jointly model all motion, often producing ghosting from transient occluders. Columns 4–5 are standard distractor-removal methods that assume a static scene and therefore struggle with valid dynamic content. ReCan-GS explicitly decomposes inconsistent observations into a distractor layer and recovers a clean, stable core reconstruction.
}
\label{fig:dyn_compare}
\label{fig:dist_compare}
\end{figure*}

\begin{figure*}[t]
\centering
\includegraphics[width=0.81\linewidth]{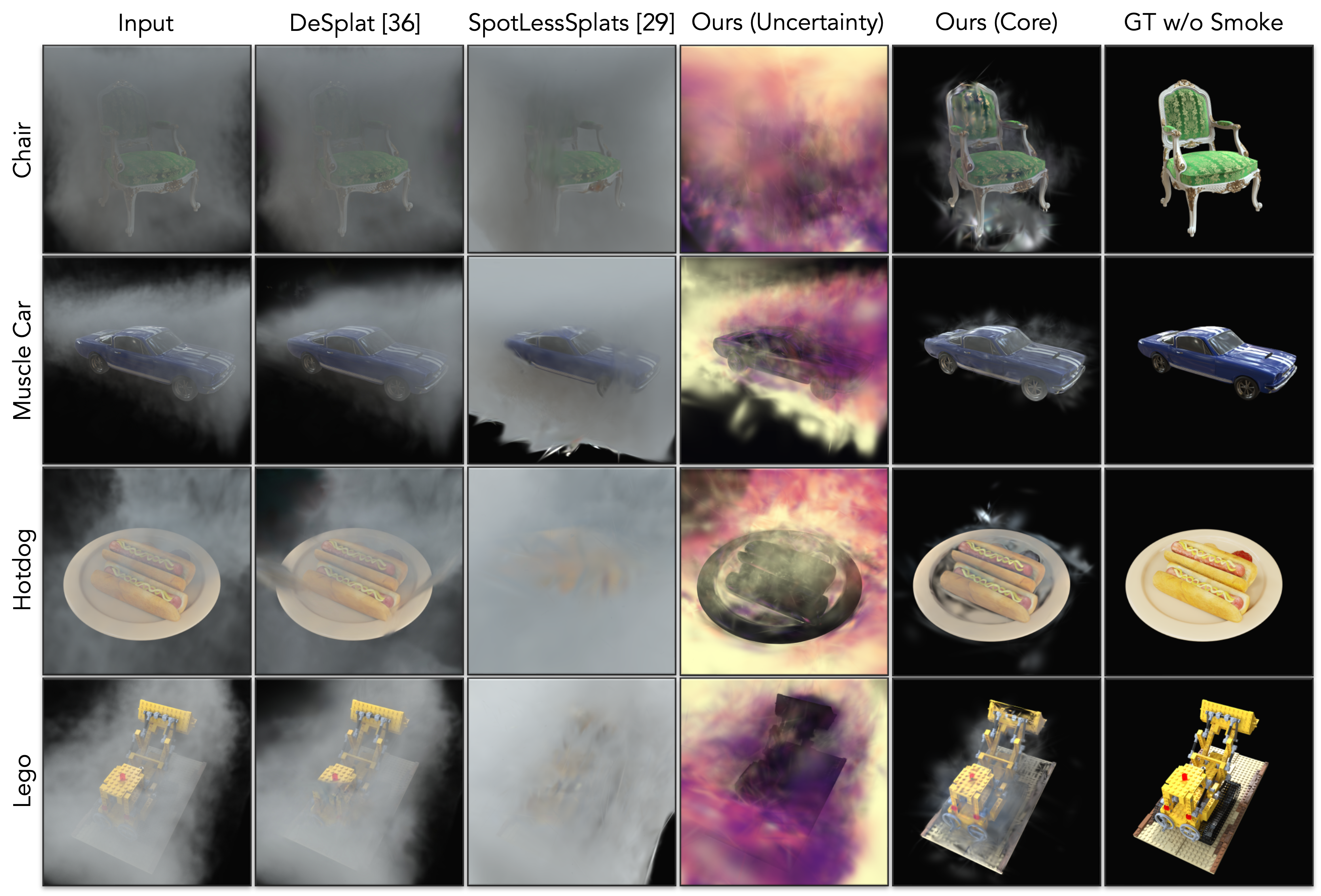}
\caption{\textbf{Smoke removal results.} ReCan-GS separates slow-moving volumetric distractors from the core scene by detecting deformation-space instability, enabling clean smoke-free reconstructions where appearance-based methods leave residual haze.
}
\label{fig:smoke_results}
\end{figure*}
To validate ReCan-GS, we evaluate across three distractor regimes that stress qualitatively different aspects of the method.

\subsection{Experimental Setup}

\paragraph{Datasets.}
Dynamic scenes test robustness to spatial and photometric occlusions; smoke scenes isolate deformation instability independent of apparent motion.

\noindent\textit{Dynamic scenes with distractors.} We utilize four scenes featuring transient occluders and lighting artifacts:

\begin{itemize}
\item Synthetic: \texttt{Bouncing Balls} (three balls occluded by a passing runner; Blender~\cite{blender2024}) and \texttt{Squatting} (a subject partially occluded by a running figure; Unity~\cite{unity2024}). We apply our loss-based regularization (Variant 1) to these scenes.
\item Real-world: Two iPhone 15 Pro captures, \texttt{Jeep} (stop-motion with heavy hand occlusions and moving cast shadows) and \texttt{Bottle} (frequently occluded by a walking person). We apply our post-hoc pruning (Variant 2) to these captures.
\end{itemize}


\noindent\textit{Smoke scenes.} To explicitly test our method's response to complex topology changes with minimal rigid motion, we evaluate on four synthetic scenes from SmokeSeer~\cite{jain2025smokeseer}: \texttt{Muscle Car}, \texttt{Chair}, \texttt{Hotdog}, and \texttt{Lego}. Generated with Blender's Mantaflow simulator over standard NeRF objects~\cite{mildenhall2020nerf}, they pair unstable smoke observations with smoke-free ground truth.



\noindent\textit{Static distractor benchmark.} To test generalization to standard static distractor removal, we evaluate on the NeRF On-the-go benchmark~\cite{Ren2024NeRF}, treating all dynamic in-the-wild content as transient.

\paragraph{Baselines and Metrics.}
We compare against dynamic scene methods (DeformableGS ~\cite{yang2023deformable3dgs}, 4DGS~\cite{Wu_2024_CVPR}) and specialized static-scene distractor-removal techniques (DeSplat~\cite{wang2024desplat}, SpotLessSplats~\cite{sabourgoli2024spotlesssplats}, WildGaussians~\cite{kulhanek2024wildgaussians}, NeRF On-the-go~\cite{Ren2024NeRF}, AsymGS~\cite{li2025asymgs}). Performance is quantified via PSNR($\uparrow$) \cite{Welstead2009FractalTechniques}, SSIM($\uparrow$) \cite{wang2004image}, and LPIPS($\downarrow$) \cite{zhang2018unreasonable}.

\vspace{-0.3cm}
\begin{table*}[t]
\centering
\scriptsize
\renewcommand{\arraystretch}{1.05}
\begin{tabular}{l
S[table-format=2.1]S[table-format=1.2]S[table-format=1.2]
S[table-format=2.1]S[table-format=1.2]S[table-format=1.2]
S[table-format=2.1]S[table-format=1.2]S[table-format=1.2]
S[table-format=2.1]S[table-format=1.2]S[table-format=1.2]}
\toprule
\multirow{2}{*}{\textbf{Method}}
& \multicolumn{3}{c}{\textbf{Squatting}}
& \multicolumn{3}{c}{\textbf{Bouncing Balls}}
& \multicolumn{3}{c}{\textbf{Jeep}}
& \multicolumn{3}{c}{\textbf{Bottle}} \\
\cmidrule(lr){2-4}\cmidrule(lr){5-7}
\cmidrule(lr){8-10}\cmidrule(lr){11-13}
& {PSNR$\uparrow$} & {SSIM$\uparrow$} & {LPIPS$\downarrow$}
& {PSNR$\uparrow$} & {SSIM$\uparrow$} & {LPIPS$\downarrow$}
& {PSNR$\uparrow$} & {SSIM$\uparrow$} & {LPIPS$\downarrow$}
& {PSNR$\uparrow$} & {SSIM$\uparrow$} & {LPIPS$\downarrow$} \\
\midrule

\multicolumn{13}{l}{\textit{Static distractor-removal methods}} \\

\rowcolor{gray!8}
WildGaussians~\cite{kulhanek2024wildgaussians}
& 24.3 & .88 & .15
& {--} & {--} & {--}
& 25.4 & .90 & .18
& 18.2 & .67 & .33 \\

SLS-MLP~\cite{sabourgoli2024spotlesssplats}
& 22.9 & .87 & .15
& 15.3 & .84 & .25
& 25.8 & .90 & .17
& 17.8 & .65 & .35 \\

\rowcolor{gray!8}
Robust3DGS~\cite{ungermann2024robust}
& 24.4 & .90 & .11
& 24.6 & .97 & .07
& 26.6 & .92 & .07
& 17.3 & .67 & .30 \\

DeSplat~\cite{wang2024desplat}
& 25.0 & .89 & .12
& 14.3 & .87 & .16
& 26.5 & .89 & .17
& 17.1 & .67 & .30 \\

\rowcolor{gray!8}
AsymGS~\cite{li2025asymgs}
& 24.7 & .88 & .12
& 21.7 & .92 & .13
& 26.3 & .91 & .09
& 17.7 & .69 & .25 \\

\midrule
\multicolumn{13}{l}{\textit{Dynamic reconstruction methods}} \\

DeformableGS \cite{yang2023deformable3dgs}
& \textbf{31.5} & \textbf{.95} & \textbf{.07}
& 27.8 & .96 & .05
& \textbf{30.3} & \textbf{.94} & \textbf{.05}
& 21.6 & .75 & .26 \\

\rowcolor{gray!8}
4DGS \cite{Wu_2024_CVPR}
& 27.2 & .79 & .27
& 22.3 & .93 & .09
& 29.2 & .91 & .16
& 16.7 & .50 & .75 \\

\rowcolor{ourscol}
\textbf{ReCan-GS (Ours)}
& 29.3 & .92 & .10
& \textbf{29.4} & \textbf{.98} & \textbf{.03}
& 28.4 & .93 & .07
& \textbf{23.0} & \textbf{.78} & \textbf{.22} \\

\bottomrule
\end{tabular}
\caption{
Quantitative results on dynamic scenes with distractors, grouped into static
distractor-removal and dynamic reconstruction methods. Higher is better for
PSNR/SSIM and lower for LPIPS; best results per scene are in \textbf{bold}. Real-scene holdout frames are distractor-free, so these metrics do not fully measure
retained distractor geometry; see Table~\ref{tab:vlm_leakage}.
}
\label{tab}
\end{table*}

\paragraph{Evaluation Protocol and Implementation.}
For synthetic scenes, metrics are computed against clean full-sequence
re-renders. For real scenes, we use distractor-free holdout subsets of 30--58
frames per scene, selected manually by one binary criterion (the distractor is
fully absent) and shared identically across all methods; frame lists are
released with the code. For the distractor-retention evaluation, we query three VLMs (GPT-5, Claude Sonnet 4.5, Gemini 2.5 Pro) with the input frame and the Core render, without indicating which method produced it, and ask for a 1–5 severity rating. Three human annotators independently rate 50 of these frames per method using the same rubric and image pairs. Distractor-present frames are determined from SAM masks independently of any method. The prompt is on the project website. Camera poses are obtained with COLMAP \cite{schoenberger2016sfm}. We train 40k iterations on synthetic scenes and 60k on
real scenes on a single NVIDIA H200, with $K=1000$ (synthetic) and $K=2000$
(real) distractor Gaussians per view, per-view densification after $S=10$
visits, and post-hoc pruning of the top $5-10\%$ most uncertain Core Gaussians. We begin deformation optimization at 3k and distractor optimization at 7k iterations, use \(\lambda_s=0.01\) and \(\lambda_{\text{unc}}=0.3\), warm \(\lambda_d\) from 0 to 0.01 over the next 10k iterations, and set \(\beta=0.9\) for post-hoc pruning. Across three random seeds, PSNR standard deviation is 0.53 dB on \textit{Bottle} and 0.09 dB on \textit{Bouncing Balls}. 



\subsection{Results}
\label{sec:results}
\paragraph{Qualitative Results.}
Fig.~\ref{fig:dyn_compare} compares against general dynamic methods
(DeformableGS~\cite{yang2023deformable3dgs}, 4DGS~\cite{Wu_2024_CVPR}) and
static distractor-removal techniques (DeSplat~\cite{wang2024desplat},
SpotLessSplats~\cite{sabourgoli2024spotlesssplats}). Dynamic methods jointly
model all motion and produce blurred, semi-transparent ghosts where transient
hands occlude the scene, while static methods damage the fidelity of the primary
moving subject. ReCan-GS isolates transient elements into the distractor layer,
leaving the core reconstruction clean and stable.

\vspace{-0.3cm}

\paragraph{Quantitative Results.}
We evaluate ReCan-GS on its primary task: reconstructing clean dynamics amidst
distractors. Table~\ref{tab} compares against both static distractor-removal
methods and general dynamic reconstruction methods. ReCan-GS substantially
outperforms the static baselines across the dynamic-distractor scenes, remains
competitive with DeformableGS in overall reconstruction fidelity, and
outperforms 4DGS on all but one scene. On \textit{Bouncing Balls}, ReCan-GS
leads all methods at 29.4 dB, while on \textit{Bottle} it achieves the best
dynamic-method result at 23.0 dB. DeformableGS leads on \textit{Squatting} and
\textit{Jeep}, and \textit{Bottle} remains our hardest scene in absolute terms
(SSIM 0.78, LPIPS 0.22). Holdout frames are selected precisely because the
distractor is absent, so a method that absorbs the occluder into its deformation
field may not be penalized for that geometry when it is absent from the evaluated renders. To assess retained distractors, we ask three human annotators
and three vision-language judges, blind to method identity, to rate per frame
how strongly the distractor is reproduced (Table~\ref{tab:vlm_leakage}). On
\textit{Jeep}, where DeformableGS leads on PSNR, they detect the occluder in
73--83\% of its frames and 77--88\% of 4DGS frames, versus 0--11\% for
ReCan-GS; no human annotator rated any ReCan-GS frame as showing the occluder.
On \textit{Bottle}, the corresponding rates are 57--77\% for DeformableGS versus 6--26\% for ReCan-GS. 4DGS leaks heavily despite the
weakest reconstruction quality, indicating that retention and reconstruction fidelity capture different aspects of performance. Relative to DeformableGS, ReCan-GS loses 2.2 dB on \textit{Squatting} and 1.9 dB on \textit{Jeep}, while gaining 1.6 dB on \textit{Bouncing Balls} and 1.4 dB on \textit{Bottle}. Lower distractor retention therefore does not imply uniformly better fidelity. Real-scene retention ratings do not verify target completeness during occlusion; removing valid target content can also reduce apparent distractors.

ReCan-GS also performs strongly on volumetric distractor removal (Table~\ref{tab:fog}, Fig.~\ref{fig:smoke_results}). While appearance-based methods often retain
semi-transparent smoke as part of the scene, ReCan-GS exploits geometric
multi-view instability to remove it. This yields gains of 5.7 dB and 6.2 dB over
the strongest baselines on the \textit{Muscle Car} and \textit{Hotdog} scenes,
respectively. These comparisons demonstrate benefits over the evaluated general distractor-removal methods; dedicated smoke-removal baselines are not included.

\vspace{-0.3cm}

\begin{table}[t]
\centering
\setlength{\tabcolsep}{4pt}
\scriptsize
\begin{tabular}{llcccc}
\toprule
& & & \multicolumn{3}{c}{VLM judges} \\
\cmidrule(lr){4-6}
Scene & Method & Human & GPT-5 & Claude & Gemini \\
\midrule
\multirow{3}{*}{\emph{Bottle}}
 & 4DGS~\cite{Wu_2024_CVPR}          & 3.41 (82.0) & 2.09 (35.5) & 3.15 (70.0) & 3.12 (75.1) \\
 & DeformableGS~\cite{yang2023deformable3dgs}  & 3.37 (76.7) & 2.60 (56.7) & 3.23 (73.8) & 3.10 (70.4) \\
 & ReCan-GS (Ours)         & \textbf{1.33 (9.3)} & \textbf{1.21 (5.6)} & \textbf{1.81 (26.0)} & \textbf{1.71 (24.7)} \\
\midrule
\multirow{3}{*}{\emph{Jeep}}
 & 4DGS~\cite{Wu_2024_CVPR}          & 3.43 (76.7) & 3.42 (82.8) & 3.68 (83.2) & 3.63 (87.8) \\
 & DeformableGS~\cite{yang2023deformable3dgs}  & 3.29 (72.7) & 3.38 (78.6) & 3.87 (78.8) & 3.46 (82.6) \\
 & ReCan-GS (Ours)         & \textbf{1.09 (0.0)} & \textbf{1.02 (0.9)} & \textbf{1.06 (1.7)} & \textbf{1.29 (11.0)} \\
\bottomrule
\end{tabular}
\caption{\textbf{Retained distractor geometry.} Three human annotators and three
VLM judges rate per frame how strongly the distractor is reproduced ($1$ =
absent, $5$ = fully reproduced); mean severity and (\% of frames rated $\ge 3$).
VLMs score all distractor-present frames (\emph{Bottle}: 534, \emph{Jeep}: 345);
the Human column averages three annotators over the same 50 frames per method,
who agree closely (mean pairwise quadratic-weighted $\kappa = 0.85$, Spearman
$\rho = 0.87$ over 900 paired ratings). On \emph{Jeep}, no annotator rated any
ReCan-GS frame $\ge 3$. Per-annotator scores are on the project website.}
\label{tab:vlm_leakage}
\end{table}

\begin{figure}[t]
    \centering
    \includegraphics[width=\linewidth]{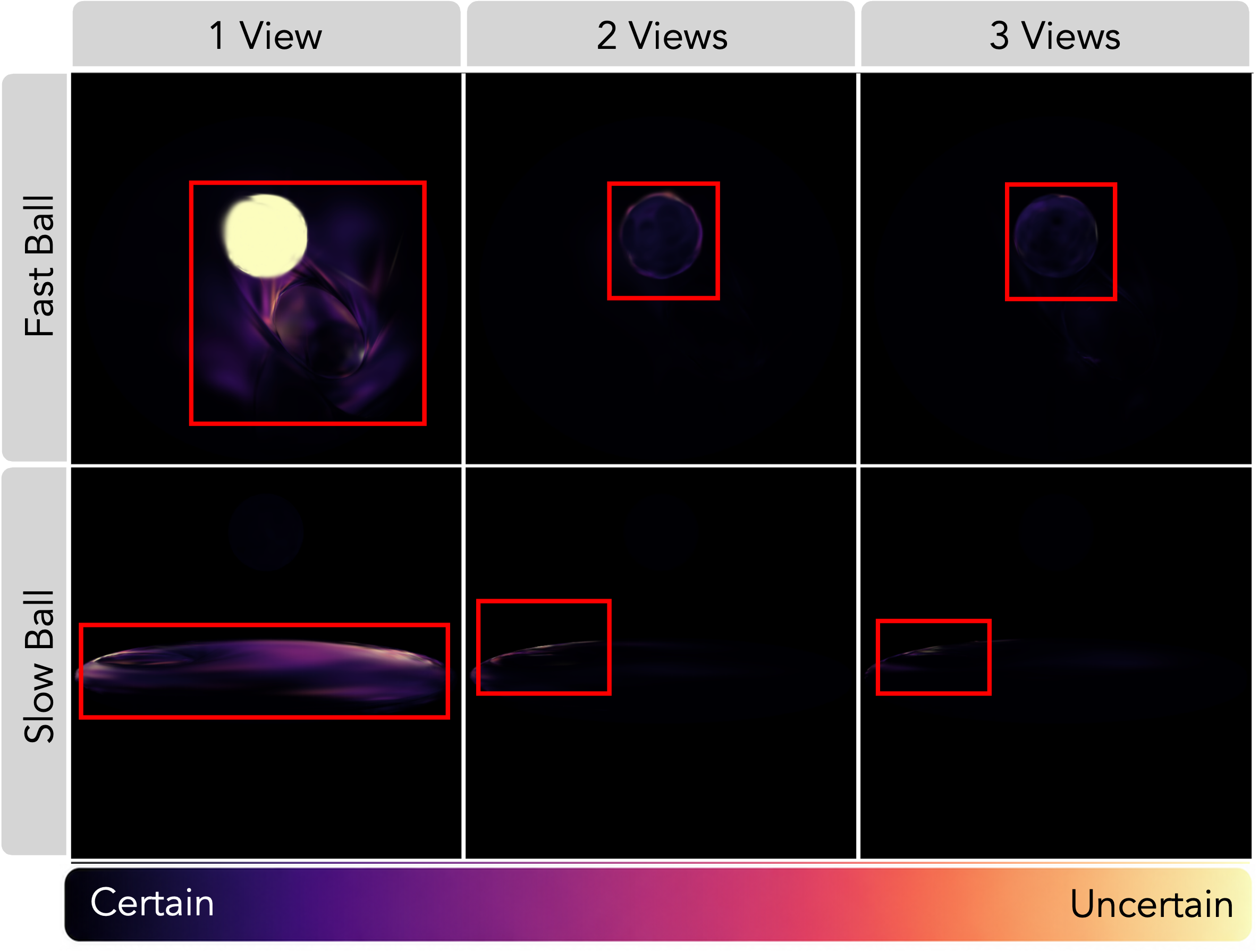}
    \caption{\textbf{Uncertainty vs. Motion and Coverage.} Uncertainty decreases as the number of camera views increases, indicating that the metric captures reconstructability rather than raw motion magnitude.}
    \label{fig:ablation}
\end{figure}

\paragraph{Computational Cost.} 
ReCan-GS increases training cost relative to 
DeformableGS (16.23 vs.\ 9.22 min; 45.9 vs.\ 72.3 
train FPS; 14.0 vs.\ 2.8 GB peak memory; 9.70M 
vs.\ 3.62M params), primarily due to the distractor 
layer and training-time Jacobian computation. It adds no inference overhead (rendering FPS is identical), and 14.0~GB fits on consumer GPUs, so the H200 is not a requirement.
\vspace{-0.3cm}

\begin{figure}[t]
    \centering
    \includegraphics[width=\linewidth]{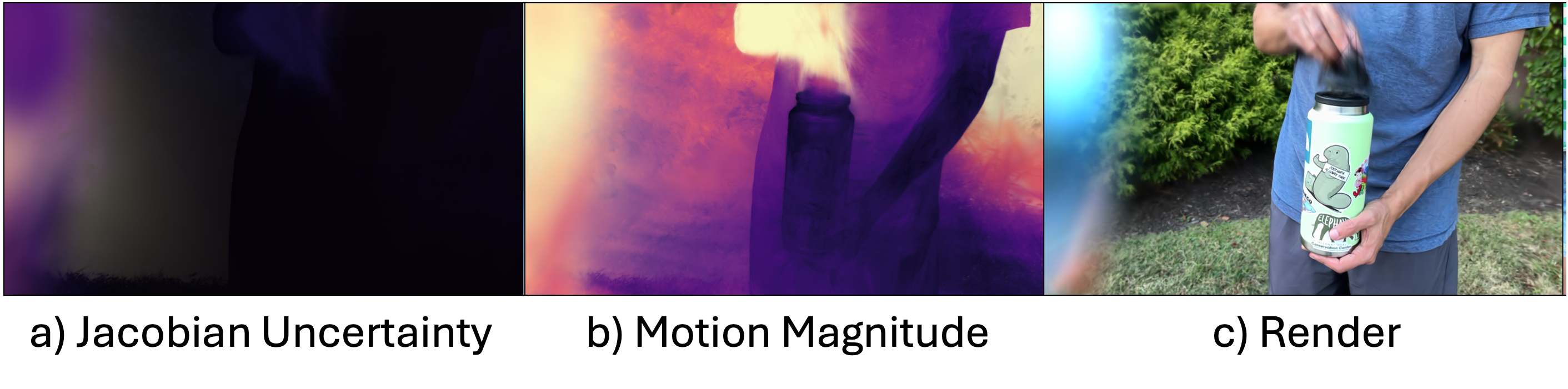}
    \caption{\textbf{Jacobian uncertainty vs.\ motion magnitude.}
Motion magnitude responds strongly to valid subject motion, whereas the
Jacobian score more selectively localizes the transient distractor.}
    \label{fig:ablation_jacvmotion}
\end{figure}

\paragraph{Spatial Localization of Transient Distractors.}
We evaluate the projected spatial support of Jacobian uncertainty on Bottle against SAM-derived masks, used only for evaluation and playing no role in the method. Across 534
distractor-containing frames, Jacobian uncertainty achieves 0.532 aggregate IoU
and 0.483 median IoU, versus 0.284 and 0.071 for motion magnitude. As shown in
Fig.~\ref{fig:ablation_jacvmotion}, motion magnitude also responds to valid
subject motion, whereas Jacobian uncertainty more selectively localizes the
transient distractor. This comparison isolates the localization signal on one scene, but does not isolate the end-to-end benefit of Jacobian routing within the same Core--Distractor architecture.

\vspace{-0.2cm}


\begin{table*}[t]
\centering
\scriptsize
\renewcommand{\arraystretch}{1.05}
\begin{tabular}{l
S[table-format=2.1]
S[table-format=1.2]
S[table-format=1.2]
S[table-format=2.1]
S[table-format=1.2]
S[table-format=1.2]
S[table-format=2.1]
S[table-format=1.2]
S[table-format=1.2]
S[table-format=2.1]
S[table-format=1.2]
S[table-format=1.2]}
\toprule
\multirow{2}{*}{\textbf{Method}}
& \multicolumn{3}{c}{\textbf{Muscle Car}}
& \multicolumn{3}{c}{\textbf{Chair}}
& \multicolumn{3}{c}{\textbf{Hotdog}}
& \multicolumn{3}{c}{\textbf{Lego}} \tabularnewline
\cmidrule(lr){2-4}
\cmidrule(lr){5-7}
\cmidrule(lr){8-10}
\cmidrule(lr){11-13}
& {PSNR$\uparrow$} & {SSIM$\uparrow$} & {LPIPS$\downarrow$}
& {PSNR$\uparrow$} & {SSIM$\uparrow$} & {LPIPS$\downarrow$}
& {PSNR$\uparrow$} & {SSIM$\uparrow$} & {LPIPS$\downarrow$}
& {PSNR$\uparrow$} & {SSIM$\uparrow$} & {LPIPS$\downarrow$} \tabularnewline
\midrule
\rowcolor{gray!8}
SLS-MLP~\cite{sabourgoli2024spotlesssplats}
& 5.4 & .01 & .71
& 6.3 & .05 & .72
& 5.6 & .14 & .76
& 4.3 & .06 & .80 \tabularnewline

DeSplat~\cite{wang2024desplat}
& 11.2 & .19 & .46
& 8.6 & .05 & .63
& 8.6 & .19 & .60
& 8.8 & .14 & .59 \tabularnewline

\rowcolor{gray!8}
Robust3DGS~\cite{ungermann2024robust}
& 10.9 & .17 & .52
& 8.6 & .05 & .67
& 8.6 & .14 & .65
& 8.2 & .09 & .67 \tabularnewline

AsymGS~\cite{li2025asymgs}
& 11.1 & 0.18 & .48
& 8.7 & .06 & .65
& 8.6 & .15 & .65
& 8.5 & .09 & .66 \tabularnewline

\midrule
\rowcolor{ourscol}
\textbf{Ours}
& \bfseries \textbf{16.8} & \bfseries \textbf{.69} & \bfseries \textbf{.26}
& \bfseries \textbf{10.9} & \bfseries \textbf{.33} & \bfseries \textbf{.50}
& \bfseries \textbf{14.8} & \bfseries \textbf{.69} & \bfseries \textbf{.30}
& \bfseries \textbf{12.5} & \bfseries \textbf{.58} & \bfseries \textbf{.39} \tabularnewline
\bottomrule
\end{tabular}
\caption{Fog/smoke removal on synthetic scenes with fog-free ground truth.
Appearance-based methods misidentify volumetric fog as stable background;
ReCan-GS removes it without any fog-specific design.
Best results in \textbf{bold}.}
\label{tab:fog}
\end{table*}

\subsection{Ablation Studies and Analysis}
\label{sec:ablation}

\paragraph{Generalization to Static Distractor Settings.}
Although ReCan-GS targets dynamic scenes, we verify it remains functional on
standard static distractor benchmarks. Table~\ref{tab:ReCan-GS_real_static}
reports results on NeRF On-the-go~\cite{Ren2024NeRF}, averaged over four scenes
(Mountain, Fountain, Corner, Patio) with all motion treated as transient.
ReCan-GS trails SLS-MLP by 2.6 dB PSNR, 0.15 SSIM, and 0.18 LPIPS. Its PSNR exceeds Splatfacto-W-A, but its SSIM and LPIPS are the weakest in this comparison. These results limit its applicability to static distractor removal; the present experiments do not isolate the cause of this gap.


\begin{table}[h]
\centering
\scriptsize
\renewcommand{\arraystretch}{1.05}
\begin{tabularx}{\columnwidth}{Xrrr}
\toprule
\textbf{Method} & \textbf{PSNR$\uparrow$} & \textbf{SSIM$\uparrow$} & \textbf{LPIPS$\downarrow$} \\
\midrule
\rowcolor{gray!8}
NeRF On-the-go~\cite{Ren2024NeRF}   & 21.32 & 0.70 & 0.25 \\
Splatfacto~\cite{nerfstudio}        & 21.28 & 0.76 & 0.15 \\
\rowcolor{gray!8}
Splatfacto-W-A~\cite{xu2024splatfactownerfstudioimplementationgaussian} & 20.54 & 0.72 & 0.18 \\
Splatfacto-W-T~\cite{xu2024splatfactownerfstudioimplementationgaussian} & 21.22 & 0.74 & 0.16 \\
\rowcolor{gray!8}
SLS-MLP~\cite{sabourgoli2024spotlesssplats} & \textbf{23.50} & \textbf{0.83} & \textbf{0.13} \\
WildGaussians~\cite{kulhanek2024wildgaussians} & 21.71 & 0.73 & 0.21 \\
\rowcolor{gray!8}
DeSplat~\cite{wang2024desplat}      & 21.70 & 0.77 & 0.14 \\
\midrule
\rowcolor{ourscol}
\textbf{Ours}                        & 20.92 & 0.68 & 0.31 \\
\bottomrule
\end{tabularx}
\caption{Static distractor generalization on NeRF On-the-go~\cite{Ren2024NeRF}.}
\label{tab:ReCan-GS_real_static}
\vspace{-0.6cm}

\end{table}

\paragraph{Validation of the Jacobian Uncertainty Metric.} To verify our metric captures geometric ambiguity rather than raw velocity, we ablate camera coverage versus motion speed (Fig.~\ref{fig:ablation}). Across fixed speeds (``Slow" and ``Fast"), increasing multi-view coverage from 1 to 3 views consistently drops the uncertainty score. This supports the score as a coverage-sensitive cue in this setting, rather than establishing a general separation of valid motion and distractors.
\vspace{-0.3cm}

\begin{figure}[h]
    \centering
    \includegraphics[width=\linewidth]{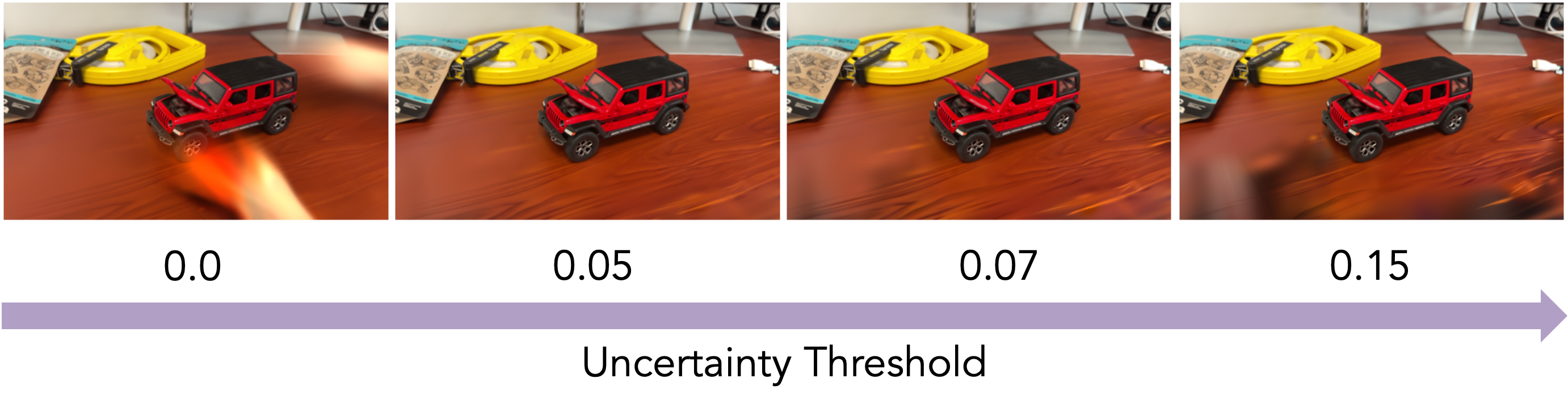}
    \vspace{-0.3cm}
    \caption{\textbf{Hierarchical Sensitivity.}
Post-hoc pruning removes the top 0\%, 5\%, 7\%, and 15\%
most uncertain Core Gaussians, ranked by accumulated
deformation sensitivity. Without pruning, transient hands
and moving cast shadows remain. Removing the top 5\%
primarily suppresses the hands, while pruning the top 7\%
and 15\% additionally suppresses moving cast shadows.}
    \label{fig:post_hoc}
\vspace{-0.3cm}
    
\end{figure}

\paragraph{Hierarchical Sensitivity of the Uncertainty Metric.}
Sweeping the post-hoc pruning fraction across the Jeep sequence
(Fig.~\ref{fig:post_hoc}) illustrates how deformation sensitivity
ranks different artifacts. Removing the highest-scoring 5\% of
Core Gaussians primarily suppresses transient hands. Increasing
the pruning fraction to 7\% and 15\% additionally suppresses
moving cast shadows.

\section{Conclusion}
We introduced ReCan-GS, a 4D Gaussian Splatting framework that uses a
Jacobian-derived deformation-sensitivity score to guide distractor separation.
Across the evaluated dynamic and synthetic smoke scenes, the same score
supports routing to a per-view Distractor layer or post-hoc pruning without
pretrained segmentation models or medium-specific supervision.
\vspace{-0.5cm}

\paragraph{Limitations and Future Work.}
Deformation sensitivity is not calibrated uncertainty or distractor identity:
stable transients may remain, while poorly observed targets or valid complex
motion may be suppressed. Retention ratings do not verify target completeness
during real occlusion. Evaluation covers four dynamic scenes (two real) and
four synthetic smoke scenes, without dedicated smoke-removal baselines or a
controlled routing ablation. Static-benchmark performance is limited
(Table~\ref{tab:ReCan-GS_real_static}). Motion or semantic priors could help
preserve intended targets.

\section*{Acknowledgments}
\begin{flushleft}
This research was supported by Fujitsu.
\end{flushleft}
{
    \small
    \bibliographystyle{ieeenat_fullname}
    \bibliography{main}
}

\end{document}